\documentclass[conference]{IEEEtran}
\IEEEoverridecommandlockouts
\usepackage{cite}
\usepackage{amsmath,amssymb,amsfonts}
\usepackage{algorithmic}
\usepackage{graphicx}
\usepackage{textcomp}
\usepackage{xcolor}
\def\BibTeX{{\rm B\kern-.05em{\sc i\kern-.025em b}\kern-.08em
    T\kern-.1667em\lower.7ex\hbox{E}\kern-.125emX}}
\begin{document}

\title{Accelerating Sparse Graph Neural Networks with Tensor Core Optimization\\}

\author{\IEEEauthorblockN{Ka Wai Wu}
\textit{Virginia Tech}\\
kawai@vt.edu \\
USA
}

\maketitle

\begin{abstract}
Graph neural networks (GNNs) have seen extensive application in domains such as social networks, bioinformatics, and recommendation systems. However, the irregularity and sparsity of graph data challenge traditional computing methods, which are insufficient to meet the performance demands of GNNs. Recent research has explored parallel acceleration using CUDA Cores and Tensor Cores, but significant challenges persist: (1) kernel fusion leads to false high utilization, failing to treat CUDA and Tensor Cores as independent resources, and (2) heterogeneous cores have distinct computation preferences, causing inefficiencies.

To address these issues, this paper proposes FTC-GNN, a novel acceleration framework that efficiently utilizes CUDA and Tensor Cores for GNN computation. FTC-GNN introduces (1) a collaborative design that enables the parallel utilization of CUDA and Tensor Cores and (2) a sparse-to-dense transformation strategy that assigns dense matrix operations to Tensor Cores while leveraging CUDA Cores for data management and sparse edge processing. This design optimizes GPU resource utilization and improves computational efficiency.

Experimental results demonstrate the effectiveness of FTC-GNN using GCN and AGNN models across various datasets. For GCN, FTC-GNN achieves speedups of 4.90×, 7.10×, and 1.17× compared to DGL, PyG, and TC-GNN, respectively. For AGNN, it achieves speedups of 5.32×, 2.92×, and 1.02×, establishing its superiority in accelerating GNN computations.

\end{abstract}


\section{Introduction}

This chapter first introduces the challenges and technical trends faced by current graph neural networks (GNNs). It then details the computational process of GNNs and associated libraries, followed by an overview of related research on accelerating GNN computation domestically and internationally. Finally, it presents the main research content and significance of this study, along with a brief outline of the paper’s structure.

\subsection{Background of the Topic}

\subsubsection{Research Background and Trends}

\begin{figure}[htbp]
    \centerline{
    \includegraphics[width=0.5\textwidth]{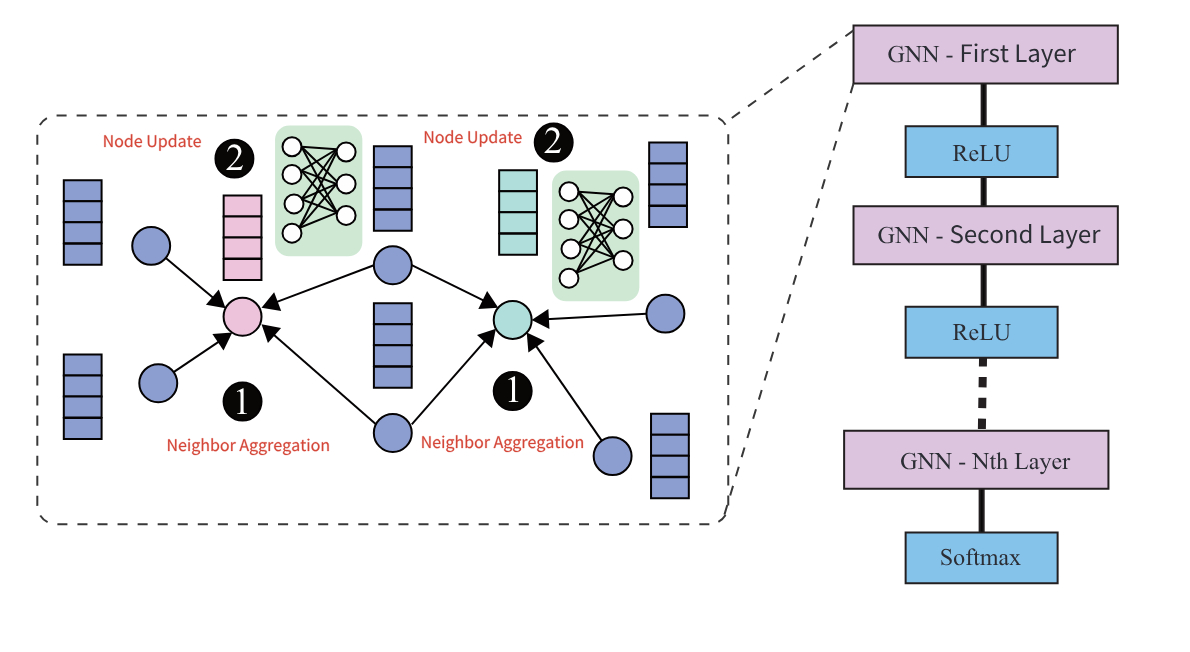}}% Replace with the path to your image file
    \caption{Computational Process of GNN}
    \label{fig:gnn_computation}
\end{figure}

GNNs have become a mainstream method for handling graph-based relational data and are widely used in fields such as security~\cite{hermes,strengthening,mitigating, Wu2025Jan,2233,2230}, data privacy~\cite{2231,2232,2234}, social network analysis~\cite{reference1, sung2025community, sung2025nlp}, pharmacology~\cite{reference3, reference4, reference5}, and recommendation systems~\cite{reference6, reference7, reference8, sung2025housing}. However, as the scale of graph data continues to grow and computational complexity increases, traditional computing methods encounter performance bottlenecks when processing large-scale GNNs. Generally, GNNs consist of a graph aggregation phase and a graph update phase. By alternating between these two phases, GNNs extract feature information from graph-structured data layer by layer, effectively representing and processing the graph data, as shown in Figure~\ref{fig:gnn_computation}.

In the aggregation phase, the main task of a GNN is to aggregate information from each node in the irregular input graph and its neighboring nodes. In the update phase, the GNN utilizes the local representations obtained from the aggregation phase to update the feature representations of each node. The basic formula for the aggregation phase is shown in Equation~\ref{eq:agg}, while the basic formula for the update phase is shown in Equation~\ref{eq:update}:

\begin{equation}
m_i^{(l)} = \text{Aggregate}(\{ h_j^{(l-1)} \mid j \in N(v_i) \})
\label{eq:agg}
\end{equation}

\begin{equation}
h_i^{(l)} = \text{Update}(m_i^{(l)})
\label{eq:update}
\end{equation}

In these equations, \( m_i^{(l)} \) in the aggregation phase represents the aggregated information for node \( v_i \) at layer \( l \), and \( h_j^{(l-1)} \) represents the feature vector of node \( v_i \)'s neighboring node \( j \) at layer \( l-1 \). In the update phase, \( h_i^{(l)} \) denotes the feature vector of node \( v_i \) at layer \( l \), while \( m_i^{(l)} \) represents the aggregated information of node \( v_i \) at the same layer.

Due to the irregular nature of graphs and memory access issues, traditional computation methods are not efficient in executing GNNs. Therefore, researchers in related fields have proposed using hardware to accelerate GNNs, such as HyGCN~\cite{reference9}. Additionally, NVIDIA introduced the Tensor Core, a new type of processing core designed specifically for matrix operations in deep neural networks, which is particularly effective for accelerating convolution and matrix multiplication operations. Tensor Core is an ideal hardware core for accelerating GNNs. TC-GNN~\cite{reference10} leverages Tensor Core to accelerate sparse GNNs by offloading parts of the GNN workload from the CUDA Core to the Tensor Core, effectively boosting overall computational speed.

\subsubsection{Aggregation and Update Phases of Graphs}

The aggregation phase in graph neural networks is responsible for iterating over each node’s neighboring nodes to gather and aggregate information from nearby nodes, thereby updating each node's features. Real-world graphs often exhibit extremely high sparsity, with adjacency matrices that can be up to 99\% sparse, making the efficient computation of sparse matrices crucial.

A sparse matrix is one in which most elements are zero or do not carry significant information. To compute on sparse matrices, sparse-matrix dense-matrix multiplication (SpMM) is commonly used to perform matrix multiplication on sparse matrices. SpMM primarily calculates only the non-zero elements of the input sparse matrix to improve the efficiency of matrix multiplication operations. The main operation involves matrix multiplication between a sparse matrix and a dense matrix to yield a dense matrix, where a dense matrix has mostly non-zero or meaningful elements. For example, given an $M \times N$ sparse matrix $A$ and an $N \times P$ dense matrix $B$, the SpMM operation calculates the product $C = A \times B$, resulting in an $M \times P$ dense matrix $C$. Compared to standard dense matrix-matrix multiplication, this approach reduces computational complexity and memory usage.

In addition to SpMM, sparse-dense-dense matrix multiplication (SDDMM) can also be used to handle sparse matrices during the aggregation phase, particularly for large sparse graphs. The primary operation in SDDMM is an element-wise multiplication between a sparse matrix and two dense matrices. For example, given an $M \times N$ sparse matrix $A$, an $M \times K$ dense matrix $B$, and an $N \times K$ dense matrix $C$, the SDDMM operation computes the element-wise product $D = A \circ (B \times C)$, where $\circ$ denotes element-wise multiplication, yielding a sparse matrix $D$. SDDMM is advantageous for improving efficiency in the aggregation phase, especially when processing large sparse graphs.

Both SpMM and SDDMM require the use of appropriate sparse matrix formats as input, such as Coordinate List (COO), Compressed Sparse Row (CSR), and Compressed Sparse Column (CSC) formats. In COO format, a sparse matrix is represented as a triplet list $(i, j, \text{value})$, where $i$ and $j$ denote the row and column indices, respectively, and \text{value} is the matrix element at that position. While COO is simple and straightforward, it records redundant information. CSR format, on the other hand, represents a sparse matrix with three arrays: a row pointer array, a column index array, and a value array. The row pointer array stores the starting index in the value array for each row, the column index array stores the column index of each non-zero element, and the value array stores the non-zero elements themselves. CSR is more memory-efficient than COO. CSC is similar to CSR but is optimized for column-access patterns. Once each node’s features have been aggregated, the aggregated features undergo a linear transformation, followed by updating the node features, leading to the update phase.

The update phase is primarily responsible for updating each node's features based on the aggregated information from its neighbors, involving dense matrix operations. Each layer in the update phase applies neural network (NN) operations such as multi-layer perceptrons (MLPs) and non-linear activation functions, with weights and bias parameters trained in this phase. As each layer in the GNN model undergoes aggregation and updating, the output of one layer serves as the input for the next, enabling the model to capture higher-order neighborhood information. The output feature embeddings, which capture structural and contextual information of nodes within the graph, can be obtained through a readout function and used for further analysis. For specific tasks such as node classification~\cite{reference12, reference13, reference14}, link prediction~\cite{reference15, reference16, reference17}, and graph classification~\cite{reference18, reference19, reference20}, the output feature embeddings are passed through a final layer, such as a fully connected layer, to generate the final output predictions.

\subsubsection{Sparse Libraries for Graph Neural Networks}

Due to the inherent sparsity of graph-structured data, GNNs often require efficient sparse matrix operations. As such, specialized sparse libraries have been designed to accelerate GNN computations, providing better performance and scalability for handling large-scale graph data.

Common GNN sparse libraries include cuSPARSE~\cite{reference21}, SuiteSparse~\cite{reference22}, and Intel MKL (Math Kernel Library)~\cite{reference23}. Intel MKL leverages multi-core CPUs and offers a range of mathematical functions and linear algebra routines, including both dense and sparse matrix operations, which can be applied to GNNs. SuiteSparse is an open-source library for computing with sparse matrices; it consists of several packages targeting different aspects of sparse matrix computations and can run on both CPU and GPU platforms. cuSPARSE is an accelerated library that provides various sparse matrix operations, including SpMM and SDDMM. However, these libraries do not leverage other computational units in the GPU beyond CUDA cores, such as the Tensor Core Unit (TCU).

The TCU is a novel computational unit developed by NVIDIA. Tensor Cores support multiple precision modes, including FP16 (half precision), TF32 (Tensor Float 32), and INT8 (8-bit integer), allowing a trade-off between computational speed and numerical precision. This approach enables accumulation of results with higher precision, reducing the risk of numerical instability. As a result, Tensor Cores can accelerate mixed-precision matrix multiplication operations. Researchers have leveraged Tensor Cores to accelerate GNN computations. For instance, Ho et al.~\cite{reference24} used Tensor Cores to accelerate general matrix multiplication (GEMM), improving overall throughput by allowing Tensor Cores to operate in parallel with CUDA Cores.

GraphSAGE, proposed by Hamilton W. and Ying Z.~\cite{reference25}, utilizes the Tensor Core’s efficiency in handling numerous simple computations to accelerate training for large-scale graphs. In a study by A. Abdelfattah, S. Tomov, and J. Dongarra~\cite{reference26}, Tensor Cores and half-precision algorithms were employed to perform small-batch matrix multiplications. Their algorithm enables faster and more energy-efficient computation for applications that can tolerate reduced numerical precision.

\section{Related Works}
\subsection{Overview}

In recent years, the application of Tensor Cores in graph neural networks (GNNs) has been widely studied both domestically and internationally, primarily in the following areas:
\begin{enumerate}
    \item \textbf{Dense Tensor Acceleration:} This is the most common method, where the sparse matrices of GNNs are converted into dense matrices, leveraging the mixed-precision computation capabilities of Tensor Cores to significantly improve the efficiency of sparse matrix multiplication.
    \item \textbf{Hybrid Computation:} This method combines the computational capabilities of Tensor Cores and CUDA Cores for more efficient computation acceleration. For example, during convolution operations, Tensor Cores handle part of the calculations, and the results are then added to the original matrix to obtain the final convolution output.
    \item \textbf{Adaptive Computational Precision:} To address potential precision errors from low-precision calculations on Tensor Cores, this method adopts different precision levels in computations to improve efficiency without affecting the final results.
\end{enumerate}

\subsection{Current State of Research}

Recent GPUs use Tensor Cores to accelerate GEMM. Since CUDA Cores remain idle during Tensor Core operations, a new parallel scheme has been proposed to split computational tasks into multiple subtasks, with some tasks offloaded from Tensor Cores to CUDA Cores. Specifically, WMMA instructions are converted into multiple MAC instructions, which are then dispatched to CUDA Cores, particularly FP32 units, to offload part of the workload from Tensor Cores. The number of MAC instructions generated depends on the matrix tile size (m-n-k) of the respective MMA instructions, and the final results are aggregated, thus enhancing overall GPU throughput. Performance bottlenecks in this offloading scheme were studied, and architectural optimizations were proposed to maximize GPU throughput. This hardware-based method does not require new compilers or other software support.

\begin{figure}[htbp]
    \centerline{
    \includegraphics[width=0.5\textwidth]{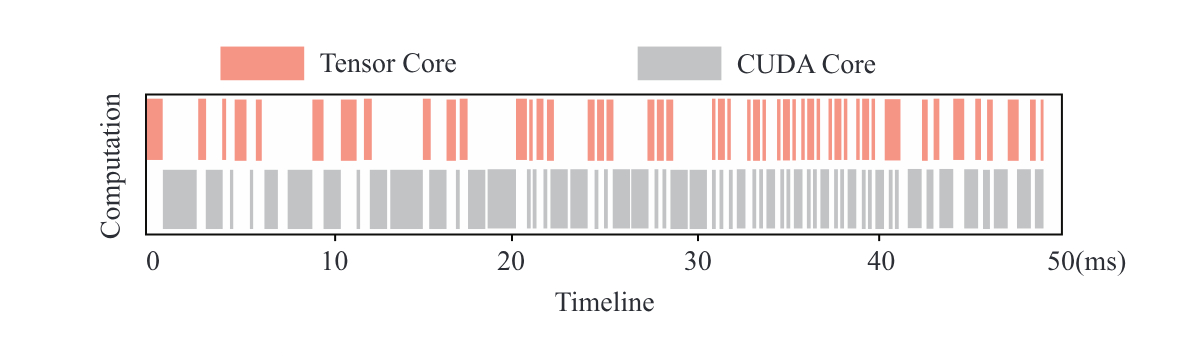} }% Replace with the path to your image file
    \caption{Active Scheduling of Tensor Cores and CUDA Cores}
    \label{fig:tacker}
\end{figure}

A framework called Tacker~\cite{reference27} was proposed, which utilizes static kernel fusion and scheduling, as shown in Figure~\ref{fig:tacker}. Tacker consists of a Tensor-CUDA kernel fuser, a fusion kernel duration predictor, and a runtime-aware QoS kernel manager. It enables simultaneous use of Tensor Cores and CUDA Cores on the GPU, enhancing GPU utilization and computational efficiency while ensuring quality of service (QoS) requirements. Tacker fuses Tensor Cores and CUDA Cores to operate in parallel. It can also merge multiple computational tasks into a single kernel, reducing memory access and data transfer for improved computational efficiency. However, if the total resource usage of two kernels exceeds the capacity of the streaming multiprocessor (SM), Tacker will not perform kernel fusion. Experimental results show that Tacker improves the throughput of backend (BE) applications by an average of 18.6\% (up to 41.1\%) while ensuring QoS.

\begin{figure}[htbp]
    \centerline{
    \includegraphics[width=0.4\textwidth]{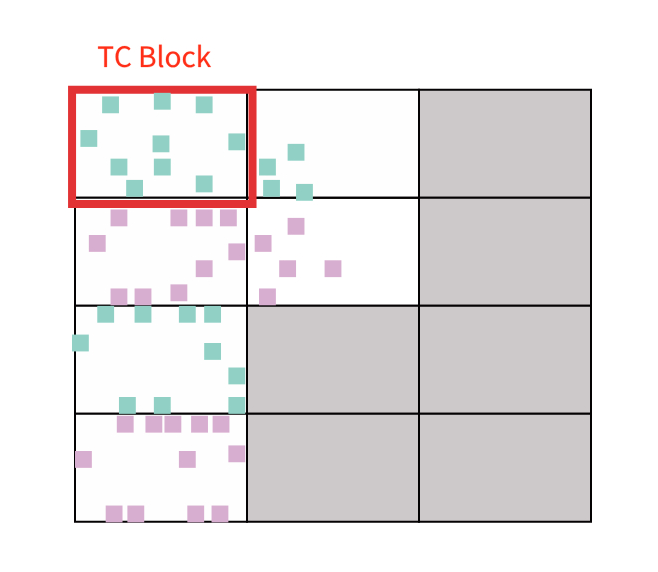}} % Replace with the path to your image file
    \caption{Dense Block Representation}
    \label{fig:dense_block}
\end{figure}

TC-GNN was proposed as a framework that uses dense tensor data to store adjacency matrices and node features, thereby achieving efficient acceleration of the entire computation process. In GNNs, graph data is usually sparse, which can impact computational speed; thus, accelerating sparse matrix computation is necessary. Libraries such as cuSPARSE and GE-SpMM~\cite{reference28} are available for sparse matrix computation, but they rely solely on the GPU's CUDA Cores, neglecting other GPU computational units like Tensor Cores. Therefore, a new sparse graph transformation technique (SGT) was introduced, which effectively combines unnecessary data loads shared by neighboring nodes into dense blocks, as shown in Figure~\ref{fig:dense_block}.

TC-GNN also implements a collaborative design between CUDA and Tensor Core Units (TCUs), mixing the CUDA Core's SIMT work mode with TCU's warp processing into a single GPU kernel. Threads from the same thread block running on CUDA Cores load data from global memory into shared memory. Once data loading is completed on CUDA Cores, each warp’s threads operate on TCUs for GEMM computation. TC-GNN is fully integrated with the PyTorch~\cite{reference29} framework, making it convenient for programming. In experiments on multiple public datasets, TC-GNN was tested and compared with existing algorithms. Results show that TC-GNN can significantly improve computational efficiency while maintaining high accuracy, achieving an average speedup of 1.7 times compared to the Deep Graph Library (DGL)~\cite{reference30} framework.

\begin{figure}[htbp]
    \centerline{
    \includegraphics[width=0.5\textwidth]{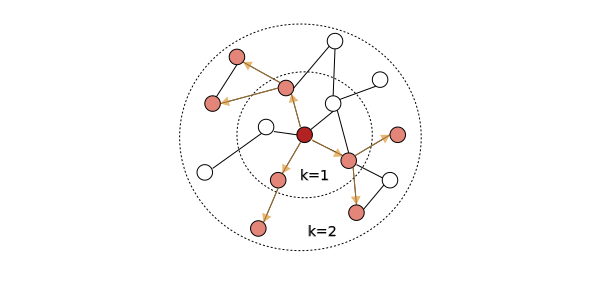}} % Replace with the path to your image file
    \caption{GraphSAGE Sampling Illustration}
    \label{fig:graphsage}
\end{figure}

GraphSAGE is an inductive learning method that generalizes to unseen graph data. Rather than training a separate embedding for each node, GraphSAGE learns a function that generates embeddings by sampling and aggregating features from a node's local neighborhood. It uses a framework similar to convolutional neural networks, enabling efficient parallel processing using computational and memory resources. GraphSAGE captures neighboring information by sampling and aggregating features, addressing memory and computational constraints of large-scale graph data. During this process, GraphSAGE samples a fixed number of nodes from a node's neighbors at each layer, aggregates their feature information, and generates the feature representation for the current node, as illustrated in Figure~\ref{fig:graphsage}.

A new graph convolutional network architecture, Spectral Graph Attention Network (SGAT), was proposed, incorporating a Fast Eigen-Approximation algorithm for graph convolution operations. SGAT, based on spectral domain methods, performs graph convolutions using the Laplacian matrix. By introducing an attention mechanism, SGAT adaptively weights each node's neighboring nodes during aggregation. Additionally, SGAT utilizes the Fast Eigen-Approximation algorithm to approximate eigenvectors and eigenvalues of the Laplacian matrix in a low-dimensional embedding space, reducing computational complexity and memory usage.

The studies above indicate a growing focus on the computational challenges of large-scale graphs and matrix multiplication. Given the computational properties of Tensor Cores, they can be introduced to accelerate operations, as exemplified by the previously mentioned GraphSAGE and SGAT. TC-GNN leverages Tensor Cores to accelerate matrix multiplication while using CUDA Cores for data management. Inspired by the parallel schemes and Tacker described above, this study proposes a new GNN acceleration framework based on TC-GNN, called FTC-GNN. FTC-GNN not only uses Tensor Cores for matrix multiplication acceleration but also offloads part of the matrix multiplication tasks to CUDA Cores to achieve even higher computational efficiency.

\section{Research Objectives and Main Content}
\subsection{Significance of the Research}

This study aims to enhance the computational performance of graph neural networks (GNNs), primarily by using Tensor Cores and CUDA Cores to improve the performance of sparse graph computations. For example, in the field of bioinformatics, GNNs are used to predict protein-protein interactions, and this acceleration design will provide researchers and practitioners with faster data processing capabilities. Since Tensor Cores are adept at handling large-scale matrix multiplications and additions and possess highly parallelized computational capabilities, they can significantly improve the efficiency of large-scale graph data processing. This has important implications for addressing issues related to massive graph data in real-world scenarios. For instance, in social network analysis, which involves handling large-scale graph data with numerous users and relationships, optimized GNNs will be able to process such data more efficiently, aiding in analyzing user behavior patterns, community structures, and other information.

\subsection{Content of the Research}

\begin{enumerate}
    \item \textbf{Implementation of Sparse Graph Transformation Technique:} This technique effectively combines unnecessary data loads shared among different nodes' neighbors. The main purpose of this algorithm is to convert sparse adjacency matrices into dense blocks and segment them into multiple TC blocks. The input to this algorithm is data in CSR format, and it compresses each row window's columns to construct TCU blocks.
    
    \item \textbf{Implementation of Sparse Neighbor Aggregation:} The sparse neighbor aggregation algorithm combines the sparse, compressed graph structure information with the dense node embedding matrix. By traversing the compressed graph and computing for each node's neighbors, it updates the node embedding matrix. This algorithm utilizes both CUDA Cores and Tensor Cores for computation.
    
    \item \textbf{Implementation of Sparse Edge Feature Computation:} The goal of edge feature computation is to calculate features for each edge in the graph. This algorithm also employs both CUDA Cores and Tensor Cores for computation.
    
    \item \textbf{Design of Collaborative CUDA and TCU Operation:} This design ensures that the sparse graph transformation technique can efficiently handle the sparse workloads of GNNs. CUDA Cores manage memory-intensive data, while TCUs handle simpler arithmetic operations.
\end{enumerate}

\subsection{Research Goals}

The primary goal of this study is to address the efficiency of sparse matrix operations in GNNs. Specifically, the research aims to design a new sparse graph transformation technique that converts sparse graphs into dense graph forms to fully leverage the computational power of Tensor Cores. Through this transformation, the study seeks to accelerate sparse matrix operations, making them suitable for the training and inference processes of GNNs.

Compared to traditional methods that solely use CUDA Cores, this study proposes a parallel acceleration scheme involving both Tensor Cores and CUDA Cores, which is expected to increase computational speed by at least 2-3 times. However, the study also aims to ensure that this acceleration approach does not compromise the accuracy of GNN models.

Accuracy is a critical factor for GNNs in practical applications, so this study must ensure that the acceleration approach achieves the expected effectiveness in real-world applications without affecting model accuracy.

\section{Design of a Sparse GNN Acceleration Scheme Based on Tensor Core}

This chapter proposes an optimized algorithm design that combines shared memory and Tensor Core Unit (TCU). First, it reviews existing hybrid sparse-dense schemes and their associated issues. Then, it explains how shared memory and TCUs can be leveraged to enhance computational performance and memory access efficiency. Finally, it introduces the GPU kernel design, hardware design, and sparse graph transformation techniques for sparse GNN computation. This approach not only fully utilizes the hardware's computational capabilities but also significantly improves computational efficiency and performance while maintaining accuracy.

\subsection{Algorithm Design Concept}

Traditional acceleration designs for sparse GNN computation typically use dense GEMM based on CUDA Cores, SpMM, and hybrid sparse-dense schemes. These methods may face certain challenges when handling graph data with varying sparsity levels.

\subsubsection{High Memory Cost}

The dense GEMM method may be more efficient for processing graph data with lower sparsity, but it can lead to additional memory consumption. Although the SpMM method is effective for sparse graph data, it may not fully utilize the GPU's computational power when sparsity is low. To make full use of the hardware's computational capability and improve performance, researchers have proposed hybrid sparse-dense schemes. This approach flexibly chooses between SpMM and GEMM operations based on the sparsity and computational requirements of the graph data.

The hybrid sparse-dense scheme combines the advantages of sparse and dense matrices to maximize computational resources and improve efficiency while maintaining accuracy. This scheme optimizes sparse computation by converting sparse computations into dense computations, addressing common issues such as irregular memory access and low computational density. When processing graph data, connections between nodes are often irregular, leading to irregular memory access during GNN computations. To address this issue, the hybrid sparse-dense scheme preprocesses the input data, reorganizing sparse data into a more manageable format.

Through compression techniques, non-zero elements are stored more compactly in memory to reduce irregular memory access. When processing sparse matrices, most elements are zero, and the distribution of non-zero elements is irregular, resulting in low computational density. To address low computational density, the hybrid sparse-dense method divides the input data into sparse and dense regions, applying specialized sparse computation techniques for sparse regions and efficient dense computation techniques for dense regions.

\begin{figure}[htbp]
    \centerline{
    \includegraphics[width=0.5\textwidth]{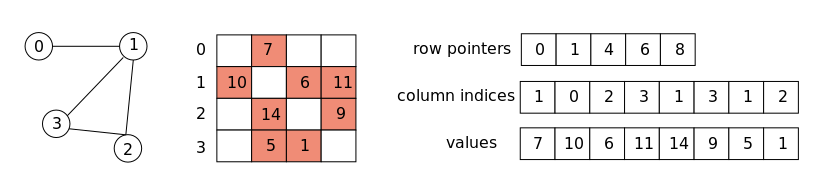}} % Replace with the path to your CSR matrix image
    \caption{CSR Representation of a Matrix}
    \label{fig:csr_matrix}
\end{figure}

The primary goal is to convert sparse matrix computations into dense matrix computations to fully leverage the hardware's computational capabilities. First, the sparse matrix is preprocessed and converted into CSR (Compressed Sparse Row) format, as illustrated in Figure~\ref{fig:csr_matrix}. Then, the CSR graph is used as input, and each row window's columns are compressed to construct TCU blocks. After that, all possible dense blocks containing non-zero elements are traversed, and GEMM operations are invoked on CUDA Cores for all confirmed non-zero blocks.

\subsubsection{Low Memory Access Efficiency}

Although the hybrid sparse-dense scheme addresses some of the aforementioned issues, it still suffers from high overhead and computational waste. First, the distribution of non-zero elements in the adjacency matrix is relatively dispersed, which makes sparse control costly, as it requires traversing all blocks to find these non-zero elements, leading to longer overall computation times.

This study introduces a novel sparse graph transformation technique to address the irregular distribution of non-zero elements in sparse graph adjacency matrices. This technique effectively identifies non-zero blocks and compresses the non-zero elements within these blocks into a smaller number of dense blocks. To this end, shared memory is utilized as a crucial space for data management within the GPU kernel, and the TCU serves as the primary computational unit. This approach allows shared memory to reorganize the irregular GNN input data in a more structured way, thereby improving memory access efficiency. Meanwhile, TCUs provide significantly higher computational throughput, further enhancing computational efficiency.

With this improved approach, this study can handle the non-zero elements in sparse graph adjacency matrices more effectively. Traditional TCU processing methods may result in a significant amount of wasted computation and memory access due to the fixed block size of TCU input matrices, which does not align with the irregular distribution of non-zero elements in sparse graphs. However, the new sparse graph transformation technique compresses non-zero elements into fewer dense blocks, allowing better utilization of TCU computation and memory access. This reduces wasted computational resources and increases block utilization, as more non-zero elements are packed into each block.

\subsubsection{Input Constraints}

Most existing GNN frameworks are built on popular NN frameworks originally optimized for dense operations. A common strategy in these frameworks is to incorporate sparse primitives into their backend implementations. However, these sparse primitives do not benefit from the same level of optimization as dense primitives and are designed to use only CUDA Cores, meaning they cannot leverage the computational power of TCUs.

\subsection{GPU Kernel Design}

This study proposes a mixed GPU kernel that combines CUDA Cores and TCUs as a single kernel, with shared memory serving as the key space for GPU data management. This design significantly enhances performance and resource utilization.

CUDA Cores operate in a Single Instruction, Multiple Threads (SIMT) manner and are managed by individual threads, while TCUs, designed for GEMM calculations, require cooperation within a warp (32 threads). Thus, kernel execution must be coordinated across different levels of execution granularity. In CUDA, threads are organized into thread blocks and grids. A thread block is a small group of threads, and thread blocks are further organized into a grid.

In this design, each row window is assigned to a thread block, with the number of threads in each block divisible by the number of threads in a warp. Data is first loaded by threads on CUDA Cores, and the matrix data is then divided into two parts: the majority is allocated to TCUs for GEMM calculations, while the remainder is handled by CUDA Cores. Specifically, TCU warp threads load data from shared memory into local registers, apply GEMM calculations to the data in registers, accumulate the results in registers, and finally store the results back to global memory. The CUDA Cores perform calculations within shared memory, and the results from both TCUs and CUDA Cores are merged.

This design has four main advantages. First, it improves memory access parallelism by distributing tasks across multiple processors or cores, shortening execution time. By processing multiple tasks simultaneously, the system can manage memory more efficiently, thereby enhancing overall performance. Second, it makes full use of memory bandwidth by balancing the load among multiple processors, ensuring each processor can maximize its available memory bandwidth.

Additionally, threads within the same warp can reuse loaded data, which avoids redundant high-cost global memory operations, reducing unnecessary memory operations and thus lowering resource consumption while improving system speed and performance. Lastly, CUDA Cores and TCUs have distinct strengths in handling different types of tasks. CUDA Cores are better suited for fine-grained thread-level execution and are ideal for managing memory-intensive data access, handling large numbers of threads simultaneously to achieve high parallelism. TCUs, on the other hand, excel at simple arithmetic operations like multiplication and addition, which is advantageous for computationally intensive tasks like GEMM operations.

Besides this design, shared memory has been customized for TCU-based sparse kernel design to accommodate intensive TCU computations, reducing redundant global memory traffic. TCU-optimized data flow designs were also integrated into neighbor aggregation (discussed in Section 3.1.2) and edge feature computation (discussed in Section 3.1.3) to achieve better performance.

\subsection{Hardware Design}

This study selects TCU as the primary computational unit over CUDA Cores. This decision is based on the significantly higher computational throughput of TCUs, which is over 10 times that of CUDA Cores. This means that when processing large volumes of data and complex mathematical operations, TCUs can complete tasks much faster than CUDA Cores. Furthermore, TCUs support high-precision calculations and allow for computations at various precision levels. This is particularly important for deep learning applications, especially GNNs, which often require handling large amounts of node and edge data, as well as executing complex mathematical operations.

However, applying TCUs directly to sparse GNN computations may lead to lower performance than using CUDA Cores. This is mainly because TCU input matrix blocks have fixed sizes, while the non-zero elements in sparse graph adjacency matrices are often distributed irregularly, resulting in excessive unnecessary computations and memory accesses.

To address this issue, this study proposes a sparse graph transformation technique. This technique effectively identifies non-zero matrix blocks and compresses the non-zero elements. By applying the sparse graph transformation technique, unnecessary data loading of shared neighbors is consolidated, thus avoiding high-cost memory access. In this way, computational efficiency is significantly improved while maintaining the accuracy of results.

The introduction of sparse graph transformation allows TCUs to better adapt to sparse GNN computations, enhancing computational performance. In sparse GNN computations, the performance gap between TCUs and CUDA Cores is narrowed, further optimizing the utilization of computational resources.

The core idea of the sparse graph transformation technique is to leverage the sparse properties of matrix blocks during sparse graph computations to reduce computation and memory access. Specifically, the technique first analyzes the input sparse graph adjacency matrix to identify non-zero elements and their corresponding matrix blocks. Next, these non-zero blocks are compressed, removing zero elements to reduce the data size. Then, by consolidating shared neighbor data loading operations, redundant loading of adjacent data is avoided, further reducing memory access overhead.

\section{Implementation of Key Techniques for Accelerating Sparse GNNs}

This chapter primarily explores three key techniques—sparse graph transformation, sparse neighbor aggregation, and sparse edge feature algorithms—that are essential for optimizing the efficiency of running GNNs on Tensor Core Units (TCUs). First, this study will provide an in-depth analysis of the sparse graph transformation technique, demonstrating how adjusting the sparsity of the graph can adapt it to the computational capabilities of TCUs. Then, it will examine the sparse neighbor aggregation algorithm, which aims to transform highly irregular SpMM into a more regular GEMM operation to take full advantage of the parallel computing capabilities of TCUs. Following this, the sparse edge feature algorithm will be introduced, along with an analysis of the similarities and differences between the sparse edge feature algorithm and the sparse neighbor aggregation algorithm, detailing the key steps in the computation process. Finally, the study explores how to use CUDA functions to interact with Python.

\subsection{Algorithm Design}

\subsubsection{Sparse Graph Transformation Technique}

This study proposes an innovative sparse graph transformation technique specifically designed to accelerate GNNs on TCUs. This technique primarily adjusts the sparsity of the graph to better align with the computational capabilities of TCUs, improving efficiency while ensuring the correctness of the output.

To achieve this goal, the sparse graph transformation preprocessing process must first be completed. This involves converting the graph’s input data structure into CSR (Compressed Sparse Row) format, where \texttt{edgeList} represents the column indices of the adjacency matrix in CSR format, and \texttt{nodePointer} represents the row pointers of the adjacency matrix in CSR format. Next, OpenMP is used to parallelize the computation process. At this step, each edge's corresponding row in the adjacency matrix is calculated based on \texttt{nodePointer}, and the neighbor ID array is sorted within each window. The results are then stored in the \texttt{edgeToRow} array.

\begin{figure}[htbp]
    \centering
    \includegraphics[width=0.5\textwidth]{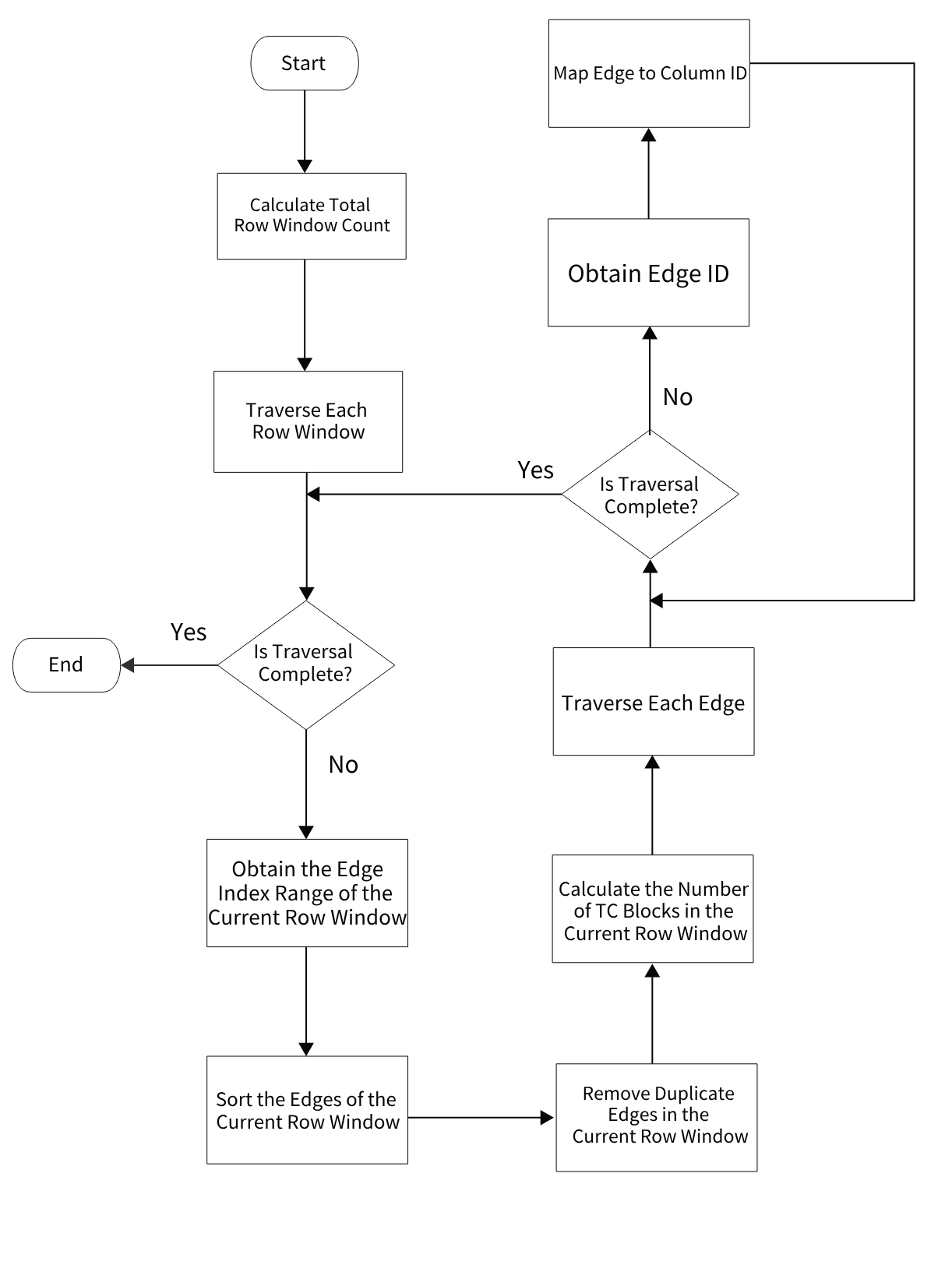} % Replace with the path to your image file
    \caption{Flowchart of Sparse Graph Transformation Technique}
    \label{fig:sparse_graph_transformation}
\end{figure}

\begin{figure}[htbp]
    \centerline{
    \includegraphics[width=0.5\textwidth]{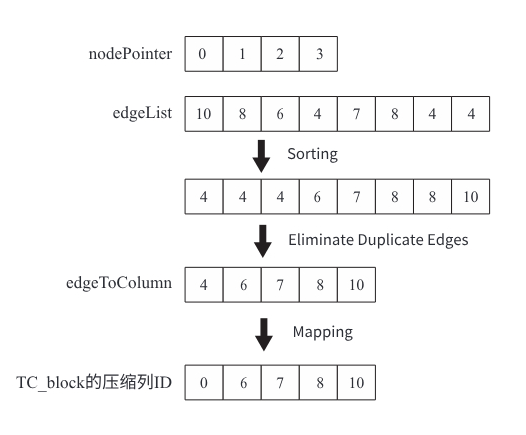}} % Replace with the path to your example image file
    \caption{Example of Sparse Graph Transformation Technique}
    \label{fig:sparse_graph_example}
\end{figure}

Then, each row window’s TCU block (TC\_block) is constructed by eliminating duplicate edges, and the total number of TC\_blocks (\texttt{block\_counter}) is updated. Additionally, the \texttt{edgeList} array is traversed to generate an edge-to-column mapping, which maps each edge ID to the compressed column ID in the TC\_block, stored in the \texttt{edgeToColumn} array to maintain the mapping relationship between edges and their corresponding positions in the compressed format. The flowchart of the sparse graph transformation technique is shown in Figure~\ref{fig:sparse_graph_transformation}, and a specific example is illustrated in Figure~\ref{fig:sparse_graph_example}.

Through this preprocessing process, the sparse graph transformation technique can more effectively adjust graph sparsity to match TCU computation. In this study, the TF-32 data format is adopted, with thread block sizes set as \texttt{TC\_BLK\_H} = 16 and \texttt{TC\_BLK\_W} = 8.

It is worth noting that the sparse graph transformation technique is not only applicable to SpMM and SDDMM operations in sparse GNNs but also has the advantage of being easily parallelizable. During sparse operations, each row window can be processed independently, which means that processing for each row window can be executed concurrently on multiple computing units (such as GPU cores or CPU threads), significantly improving computational efficiency.

Another important advantage of the sparse graph transformation technique is its one-time execution for sparse graph transformation. During this process, the original graph data structure is converted into a format more suitable for SpMM and SDDMM operations. Since this transformed format can be reused in multiple rounds of GNN training and inference, it eliminates the need to re-execute the transformation process in each iteration. This feature reduces computational burden, saves a considerable amount of time, and enhances overall performance.

\subsubsection{Sparse Neighbor Aggregation}

This study develops a sparse neighbor aggregation algorithm based on the characteristics of TCUs. The neighbor aggregation algorithm aggregates the features of neighboring nodes into the target node to generate a new feature representation. The primary objective is to transform the highly irregular SpMM operation into a more regular GEMM operation. This transformation allows the algorithm to leverage the high parallelism and high throughput capabilities of TCUs.

Specifically, this study implements a CUDA kernel to efficiently compute the matrix multiplication of a sparse matrix \( A \) and a dense matrix \( X \) on TCUs. The primary application scenario for this kernel is in GNNs, where the sparse matrix \( A \) represents the adjacency relationships between nodes, and the dense matrix \( X \) represents node features. Through this kernel, neighbor aggregation operations can be executed in parallel on the GPU, accelerating the training and inference processes of GNNs.

First, several kernel parameters are defined, including the node pointer array \texttt{nodePointer}, edge list array \texttt{edgeList}, array \texttt{blockPartition} representing the number of TC blocks per row window, edge-to-column mapping array \texttt{edgeToColumn}, edge-to-row mapping array \texttt{edgeToRow}, number of nodes \texttt{numNodes}, number of edges \texttt{numEdges}, embedding dimension \texttt{embedding\_dim}, input feature matrix \texttt{input}, and output feature matrix \texttt{output}.

The GPU kernel uses a three-level parallel strategy involving thread blocks, warps, and threads. Specifically, thread blocks handle a portion of the output matrix, while warps and threads perform matrix operations at different granularities in parallel. To cache frequently used information in on-chip shared memory—including storing the sparse matrix \( A \), the column ID mapping of the sparse matrix \( A \) to the row ID of the node embedding matrix \( X \) (\texttt{sparse\_AToX\_index}), and the dense matrix \( X \)—the \texttt{\_\_shared\_\_} modifier is used to enable data reuse. Additionally, half-precision floating-point numbers are employed to reduce computational complexity and memory usage.

The TC blocks are divided into two parts, with most handled by TCUs. The main loop of the kernel processes TC blocks along the column dimension of the sparse matrix \( A \). In each iteration, \texttt{BLK\_H} threads from \texttt{warp-0} first initialize the sparse matrix \( A \). Then, threads in a warp initialize the dense matrix \( X \). Next, the \texttt{load\_matrix\_sync} function from the WMMA API loads fragments of the sparse matrix \( A \) and the dense matrix \( X \) into fragment variables. The matrix multiplication operation is then performed by calling \texttt{wmma::mma\_sync}, multiplying fragments of the sparse matrix \( A \) and dense matrix \( X \), with the results accumulated into an accumulator fragment.

The remaining TC blocks are processed on CUDA Cores, where scalar multiplication and addition are performed on the non-zero elements of the matrix during each TC block traversal.

\begin{figure*}[htbp]
    \centerline{
    \includegraphics[width=0.8\textwidth]{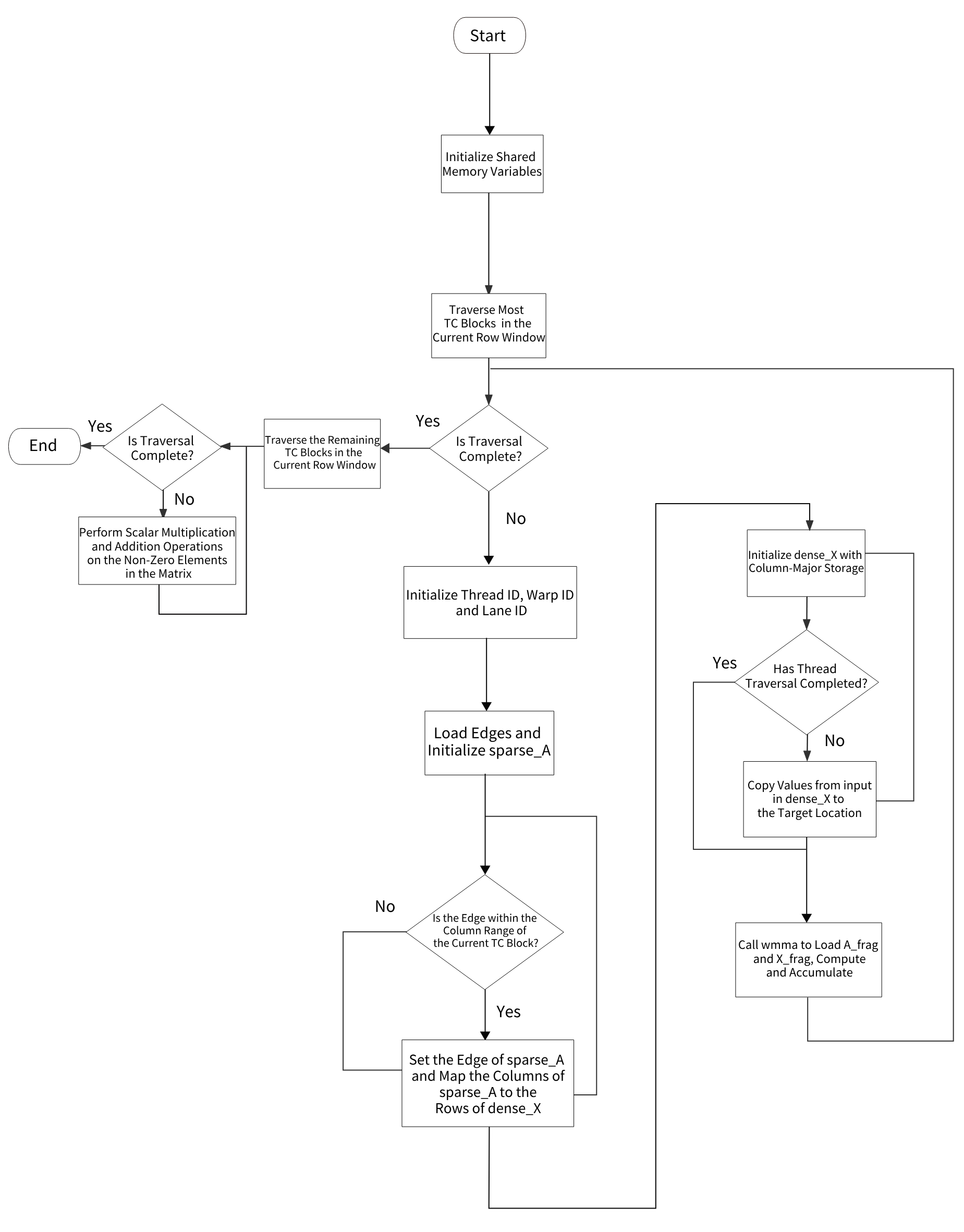} }% Replace with the path to your image file
    \caption{Flowchart of Sparse Neighbor Aggregation}
    \label{fig:sparse_neighbor_aggregation}
\end{figure*}

In this process, the \texttt{\#pragma unroll} directive is used to unroll loops, improving computational performance. After processing all TC blocks, the \texttt{wmma::store\_matrix\_sync} function stores the results from the accumulator fragment into the output matrix. The algorithm flow is shown in the flowchart in Figure~\ref{fig:sparse_neighbor_aggregation}. Note that to ensure the correctness of the output, the embedding dimension must be divisible by \texttt{BLK\_H}.

This CUDA kernel efficiently computes the multiplication of sparse and dense matrices on the GPU, accelerating the training and inference processes of GNNs. The kernel leverages the high parallelism of GPUs and Tensor Core technology, achieving high-performance processing of large-scale graph data.

In the core part of the code, the WMMA (Warp Matrix Multiply-Accumulate) API is used to perform matrix multiplication by defining fragments for matrices \( A \) and \( B \) and an accumulator fragment. During this process, the fragment for matrix \( A \) is defined in row-major order, and the fragment for matrix \( B \) is defined in column-major order. The WMMA API functions for TCUs are listed in Table~\ref{table:wmma_api}.

\begin{table*}[t]
    \centering
    \caption{WMMA API for TCUs}
    \label{table:wmma_api}
    \begin{tabular}{|c|c|}
        \hline
        Operation & Function \\
        \hline
        Declare a WMMA fragment & \texttt{wmma::fragment a\_frag;} \\
        Load data from matrix \( A \) into the WMMA fragment \texttt{a\_frag} & \texttt{wmma::load\_matrix\_sync(a\_frag, A, M);} \\
        Perform a WMMA matrix multiply-accumulate operation & \texttt{wmma::mma\_sync(c\_frag, a\_frag, b\_frag, c\_frag);} \\
        Store data from the WMMA fragment \texttt{c\_frag} into matrix \( C \) & \texttt{wmma::store\_matrix\_sync(C, c\_frag, N, mem\_row\_major);} \\
        \hline
    \end{tabular}
\end{table*}
\subsubsection{Sparse Edge Feature Computation}

The sparse edge feature computation is similar to the sparse neighbor aggregation algorithm in terms of overall algorithm structure and input, but differs in its output. Specifically, the objective of neighbor aggregation is to update node embeddings using the feature information of nodes and their neighbors. Therefore, in the neighbor aggregation process, this study focuses on each node and its neighbors, integrating this information into the updated node embedding matrix. The output is a dense node embedding matrix, which can be used in subsequent GNN layers to further extract and propagate node features.

Unlike neighbor aggregation, the goal of edge feature computation is to calculate features for each edge in the graph. In this process, the focus is on the feature information of the source and target nodes of each edge. This information is aggregated into a sparse matrix, and the final output is in a sparse format with the same shape as \texttt{edgeList}, used to store edge features. This sparse matrix can serve as input for subsequent operations, such as edge classification, prediction, and aggregation tasks.

\begin{figure*}[h]
    \centering
    \includegraphics[width=0.8\textwidth]{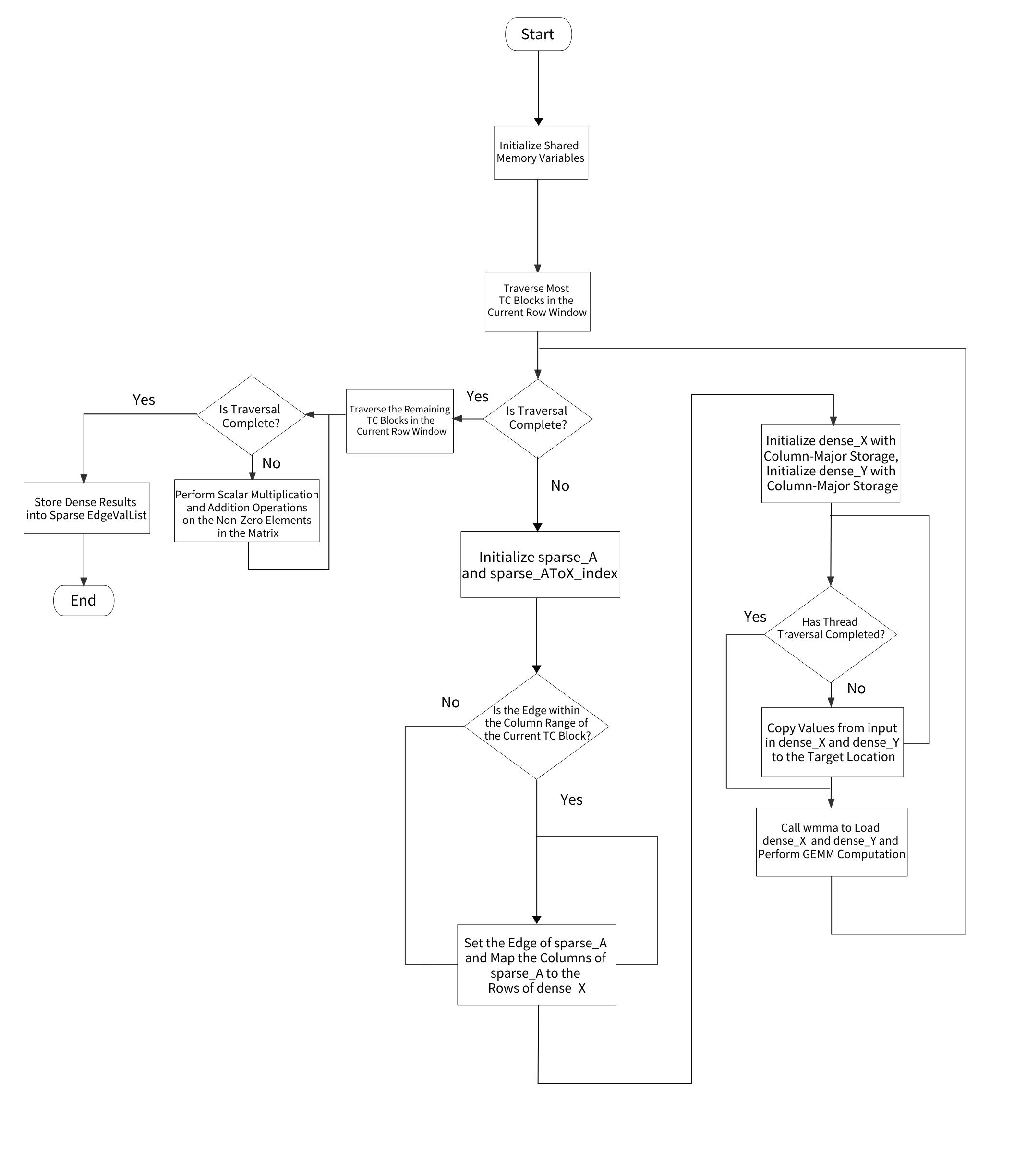} % Replace with the path to your image file
    \caption{Flowchart of Edge Feature Computation}
    \label{fig:edge_feature_computation}
\end{figure*}

The specific algorithm flow for sparse edge feature computation is shown in the flowchart in Figure~\ref{fig:edge_feature_computation}.

First, four shared memory arrays are used: the edge data of the sparse matrix \( A \) (\texttt{sparse\_A}), the column index mapping of the sparse matrix \( A \) to the row index of the dense matrix \( X \) (\texttt{sparse\_AToX\_index}), the dense input feature matrix \( X \) (\texttt{dense\_X}), and the dense input feature matrix \( Y \) (\texttt{dense\_Y}).

An outer loop is then used to handle most of the TC blocks in the current row window, and a dummy value is used to initialize the \texttt{sparse\_AToX\_index} array. Next, the sparse matrix \( A \) is initialized in parallel using the thread ID (\texttt{tid}). After initializing the sparse matrix \( A \), 16 threads from \texttt{warp-0} are used to traverse all neighbors of the current node. For each neighbor, its column index in the sparse matrix \( A \) is obtained first, and it is checked whether the neighbor is within the column range of the current TC block. If so, the edge data is stored in the sparse matrix \( A \), and the column index of the sparse matrix \( A \) is mapped to the row index of the dense matrix \( X \).

After processing all neighbors of the current TC block, all dimensions of the same sparse block are traversed. First, the dense matrix \( X \) is initialized in parallel using the thread ID. In this process, row-major storage and boundary checks are handled. Next, the dense matrix \( Y \) is initialized in parallel using the thread ID. At this stage, column-major storage and boundary checks are handled.

Once the dense matrices \( X \) and \( Y \) have been initialized, the \texttt{\_\_syncthreads()} function is used to synchronize all threads. This synchronization ensures that all threads complete the matrix initialization before proceeding to the next operation. The \texttt{wmma} library functions are then called on the GPU's Tensor Cores to perform GEMM calculations on the dense matrices \( X \) and \( Y \). This step combines information from the feature matrices \( X \) and \( Y \) to calculate edge features. The remaining TC blocks are processed on CUDA Cores, where scalar multiplication and addition operations are performed on the non-zero elements of the matrix during each TC block traversal. Finally, the results from the TCU and CUDA Core computations are merged, and the results are stored in the sparse \texttt{EdgeValList}. This completes the entire process of edge feature computation.

The shared memory design for edge feature computation is similar to that in neighbor aggregation but differs in data flow design, the size of the sparse matrix \( A \), and the number of iterations over the embedding dimension. First, in neighbor aggregation, the sparse matrix \( A \) serves as an input for an SpMM-like operation, with a minimum processing granularity of \( 16 \times 8 \). However, in edge feature computation, the sparse matrix \( A \) follows an SDDMM-like operation as the output matrix, maintaining a minimum processing granularity of \( 16 \times 16 \). Furthermore, in edge feature computation, it is necessary to recalculate the total number of TC blocks to reuse the transformed sparse graph in the SpMM operation.

Edge feature computation also requires accumulating results along the embedding dimension after all iterations are completed, so the output can only be produced after the entire iteration process concludes. In contrast, in the neighbor aggregation method, node embedding vectors are allocated to several warps, with each warp outputting its aggregated results in parallel to non-overlapping embedding dimension regions.

\section{Performance Testing and Analysis}

This chapter focuses on testing the sparse GNN computation introduced with Tensor Cores and provides a detailed analysis of the experimental results. First, the hardware and software environment used in the experiments will be introduced. Then, the design of the experimental scheme will be explained in detail. Finally, the experimental results will be analyzed, and the impact of parallel execution of Tensor Cores and CUDA Cores on the performance of sparse GNN computation will be discussed.

\subsection{Experimental Software and Hardware Environment}
\subsubsection{Server Hardware Environment}

For the software environment, suitable software tools and libraries were chosen to support sparse GNN computation. The selection of these tools and libraries was carefully considered to ensure the effectiveness and accuracy of sparse GNN computation by providing robust functionality and performance. The specific software environment is shown in Table~\ref{table:software_environment}.

\begin{table}[h]
    \centering
    \caption{Experimental Software Environment}
    \label{table:software_environment}
    \begin{tabular}{|c|c|}
        \hline
        Software & Version \\
        \hline
        Operating System & Ubuntu 18.04.3 LTS \\
        Kernel & Linux 4.15.0-208-generic \\
        PyTorch & 1.10.0 \\
        Python & 3.6.2 \\
        gcc & 7.5.0 \\
        CUDA & 11.3 \\
        cuDNN & 8.2.0 \\
        nvcc & 11.3 \\
        DGL & 0.9.1 \\
        PyTorch Geometric & 2.0.3 \\
        \hline
    \end{tabular}
\end{table}

For hardware, a remote server equipped with a high-performance GPU was used as the experimental platform. This server has ample computational power and memory resources to meet the requirements of sparse GNN computation in this study. The detailed hardware environment is shown in Table~\ref{table:hardware_environment}.

\begin{table}[h]
    \centering
    \caption{Experimental Hardware Environment}
    \label{table:hardware_environment}
    \begin{tabular}{|c|c|c|}
        \hline
        Hardware Device & Model & Quantity \\
        \hline
        GPU & NVIDIA Tesla V100 & 1 \\
        CPU & Intel Xeon Processor (Skylake, IBRS) & 2 \\
        Memory & 32GB & 1 \\
        Graphics Card & Cirrus Logic GD 5446 & 1 \\
        \hline
    \end{tabular}
\end{table}
\subsubsection{Testing Tools}

PyTorch, launched by the Facebook AI Research team in 2016, has become one of the most popular frameworks in the fields of machine learning and deep learning. It is a Python-based scientific computing library mainly used for building deep learning models. In this study, PyTorch was used to write code for accelerating sparse GNNs, as well as for training and testing the model using the tools and functions provided by PyTorch. PyTorch offers a wealth of functionalities and a flexible interface, allowing for easy definition and optimization of GNN models. By leveraging the powerful capabilities of PyTorch, the study was able to efficiently implement and evaluate the model, validating its performance.

\subsection{Experimental Design and Result Analysis}

In this study, several datasets were selected for testing, including \texttt{citeseer}, \texttt{cora}, \texttt{amazon0505}, \texttt{com-amazon}, and \texttt{amazon0601}. The specific number of TC blocks and edges in each dataset is shown in Table~\ref{table:dataset_stats}. These datasets have wide-ranging applications across different fields and cover various types of graph structures, making the experimental results more generalizable.

\begin{table}[h]
    \centering
    \caption{Number of TC Blocks and Edges in the Datasets}
    \label{table:dataset_stats}
    \begin{tabular}{|c|c|c|}
        \hline
        Dataset & Number of TC Blocks & Number of Edges \\
        \hline
        citeseer & 659 & 84352 \\
        cora & 681 & 87168 \\
        amazon0505 & 234206 & 29978368 \\
        com-amazon & 124398 & 15922944 \\
        amazon0601 & 164737 & 21086336 \\
        \hline
    \end{tabular}
\end{table}

To investigate the performance of GNNs on these datasets, this study selected the Graph Convolutional Network (GCN) and Attention-based Graph Neural Networks (AGNN) as experimental models.

GCN is a semi-supervised learning method specifically designed for handling graph-structured data. Its core concept is to perform convolution operations within the local neighborhood of nodes, capturing information between nodes and their neighbors. By performing convolution operations on graphs, GCN can effectively propagate and aggregate node features, thereby capturing relationships among nodes in the graph. This capability allows GCN to perform excellently in tasks such as node classification, link prediction, and graph classification. In the experiments, a 2-layer GCN with 16 hidden dimensions was used, matching the configuration of TC-GNN.

AGNN is an attention-based GNN that selectively aggregates information from neighboring nodes and edges, taking into account their relevance to the task. This attention mechanism enables AGNN to capture both local and global dependencies in the graph, making it useful for tasks such as node classification, link prediction, and graph classification. In the experiments, a 4-layer AGNN with 32 hidden dimensions was used, also matching the configuration of TC-GNN. By experimenting with these two models, this study explores their performance on the selected datasets.
\subsubsection{Comparison with DGL}

DGL is an open-source Python library for GNNs that provides an efficient, flexible, and easy-to-use approach for implementing complex graph learning algorithms. DGL aims to simplify the implementation of GNN models and supports various graph structures, including directed, undirected, heterogeneous, and hypergraphs. By using heterogeneous graph data representation, DGL can better handle graphs with multiple node and edge types. DGL offers efficient implementations of GNN algorithms and supports parallel computing and distributed training, enabling efficient model training on large-scale graph data. It also supports GPU acceleration, leveraging GPU parallel computing capabilities to speed up model training. Additionally, DGL provides rich graph operations and feature engineering tools, making it easy to preprocess and extract features from graph data. Through these features, DGL has become an essential tool for researchers and developers in the field of GNNs, advancing graph data analysis and graph machine learning.

\begin{figure}[htbp]
    \centerline{
    \includegraphics[width=0.5\textwidth]{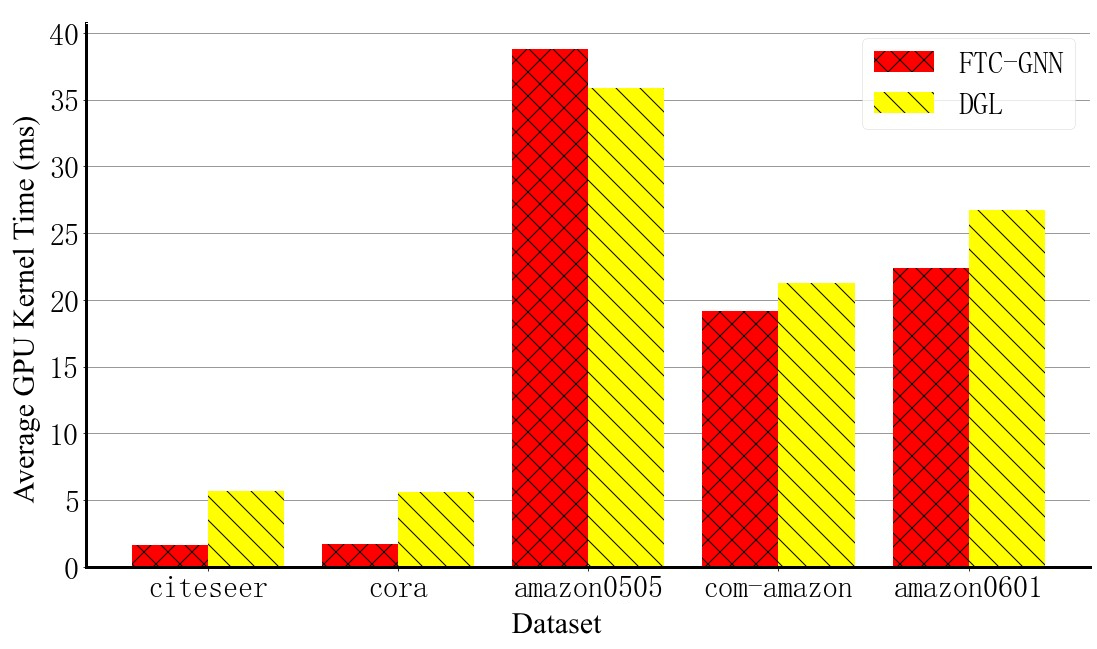}} % Replace with your actual image path
    \caption{Comparison of Average GPU Kernel Time Between FTC-GNN and DGL (GCN)}
    \label{fig:ftc_dgl_gcn}
\end{figure}

\begin{figure}[htbp]
    \centerline{
    \includegraphics[width=0.5\textwidth]{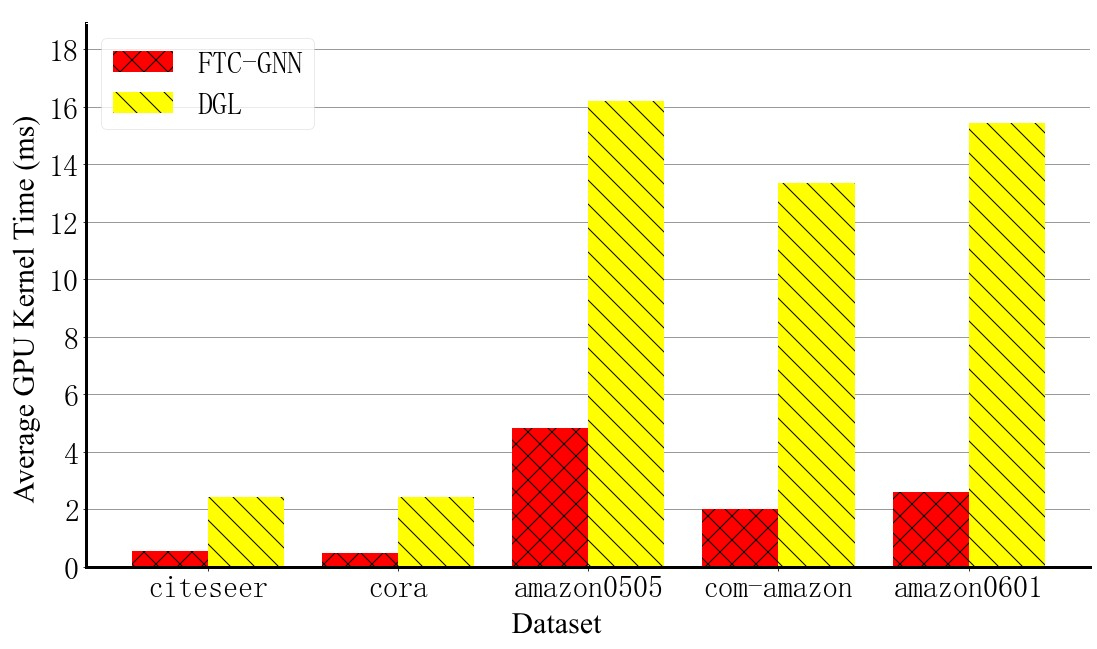}} % Replace with your actual image path
    \caption{Comparison of Average GPU Kernel Time Between FTC-GNN and DGL (AGNN)}
    \label{fig:ftc_dgl_agnn}
\end{figure}

As shown in Figures~\ref{fig:ftc_dgl_gcn} and~\ref{fig:ftc_dgl_agnn}, FTC-GNN achieved an average speedup of 4.90 times for GCN and 5.32 times for AGNN compared to DGL across multiple datasets.

This result was obtained by measuring the time required for FTC-GNN and DGL to complete the same tasks and calculating the speedup ratio. Specifically, the speedup ratio was calculated by dividing the GPU kernel runtime of DGL by the runtime of FTC-GNN, with the average taken across multiple test cases.

For GCN, the specific average GPU kernel times of FTC-GNN and DGL are shown in Table~\ref{table:ftc_dgl_gcn}. Comparing the average GPU kernel time of FTC-GNN and DGL, it is evident that FTC-GNN has shorter average GPU kernel times than DGL across all datasets. For the \texttt{citeseer} and \texttt{cora} datasets, FTC-GNN's average GPU kernel time is significantly lower than that of DGL. For the \texttt{amazon0505}, \texttt{com-amazon}, and \texttt{amazon0601} datasets, FTC-GNN's average GPU kernel time is also lower, though the difference with DGL is slightly reduced.

\begin{table}[h]
    \centering
    \caption{Average GPU Kernel Time of FTC-GNN and DGL (GCN)}
    \label{table:ftc_dgl_gcn}
    \begin{tabular}{|p{1.5cm}|p{3cm}|p{3cm}|}
        \hline
        Dataset & FTC-GNN Average GPU Kernel Time (ms) & DGL Average GPU Kernel Time (ms) \\
        \hline
        citeseer & 0.551 & 2.450 \\
        cora & 0.485 & 2.419 \\
        amazon0505 & 4.846 & 16.193 \\
        com-amazon & 2.025 & 13.338 \\
        amazon0601 & 2.604 & 15.455 \\
        \hline
    \end{tabular}
\end{table}

Overall, FTC-GNN demonstrates shorter average GPU kernel times under the GCN model, indicating an advantage in computational efficiency.

For AGNN, the specific average GPU kernel times of FTC-GNN and DGL are shown in Table~\ref{table:ftc_dgl_agnn}. Comparing the average GPU kernel times, FTC-GNN sometimes has shorter times than DGL and sometimes longer. For the \texttt{citeseer} and \texttt{cora} datasets, FTC-GNN's average GPU kernel time is slightly lower than that of DGL. However, for the \texttt{amazon0505}, \texttt{com-amazon}, and \texttt{amazon0601} datasets, FTC-GNN's average GPU kernel time is significantly higher than that of DGL.

\begin{table}[htbp]
    \centering
    \caption{Average GPU Kernel Time of FTC-GNN and DGL (AGNN)}
    \label{table:ftc_dgl_agnn}
    \begin{tabular}{|p{1.5cm}|p{3cm}|p{3cm}|}
        \hline
        Dataset & FTC-GNN Average GPU Kernel Time (ms) & DGL Average GPU Kernel Time (ms) \\
        \hline
        citeseer & 1.665 & 5.673 \\
        cora & 1.744 & 5.630 \\
        amazon0505 & 38.762 & 35.841 \\
        com-amazon & 19.179 & 21.269 \\
        amazon0601 & 22.358 & 26.754 \\
        \hline
    \end{tabular}
\end{table}

Overall, under the AGNN model, FTC-GNN and DGL show relatively similar performance in terms of average GPU kernel time, indicating no significant advantage or disadvantage in computational efficiency.

\subsubsection{Comparison with PyG}

PyG (PyTorch Geometric) is a GNN library built on PyTorch, designed to simplify the implementation and application of GNN models. PyG provides a variety of graph learning algorithms, graph generation models, graph classification models, and other useful tools, such as data loaders and data processing utilities. PyG is primarily designed for homogeneous graphs, which offer a simpler and more understandable graph data representation. Using PyG, various GNN models can be easily constructed and trained. PyG offers efficient graph operations and computation methods, making the process of handling graph data more straightforward and efficient. It also includes many classical graph learning algorithms and models.

Additionally, PyG provides various practical tools for handling graph data, including data loaders for convenient data loading and preprocessing. PyG also provides a series of functions and operations for graph data processing, such as graph partitioning and node feature transformation.

\begin{figure}[htbp]
    \centerline{
    \includegraphics[width=0.5\textwidth]{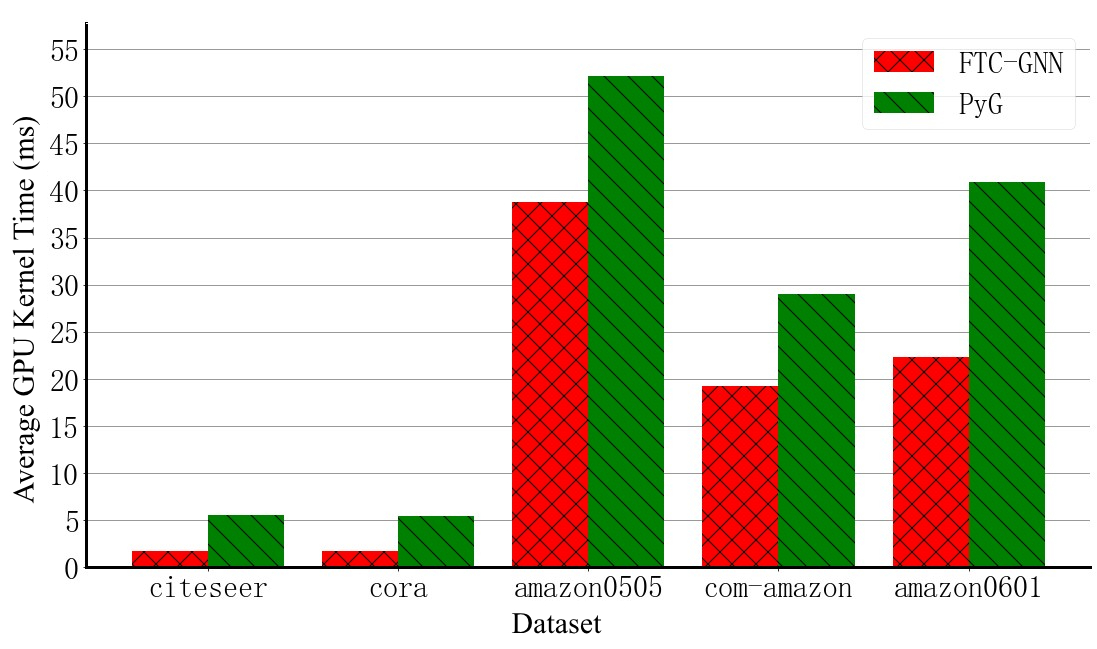}} % Replace with your actual image path
    \caption{Comparison of Average GPU Kernel Time Between FTC-GNN and PyG (GCN)}
    \label{fig:ftc_pyg_gcn}
\end{figure}

\begin{figure}[htbp]
    \centerline{
    \includegraphics[width=0.5\textwidth]{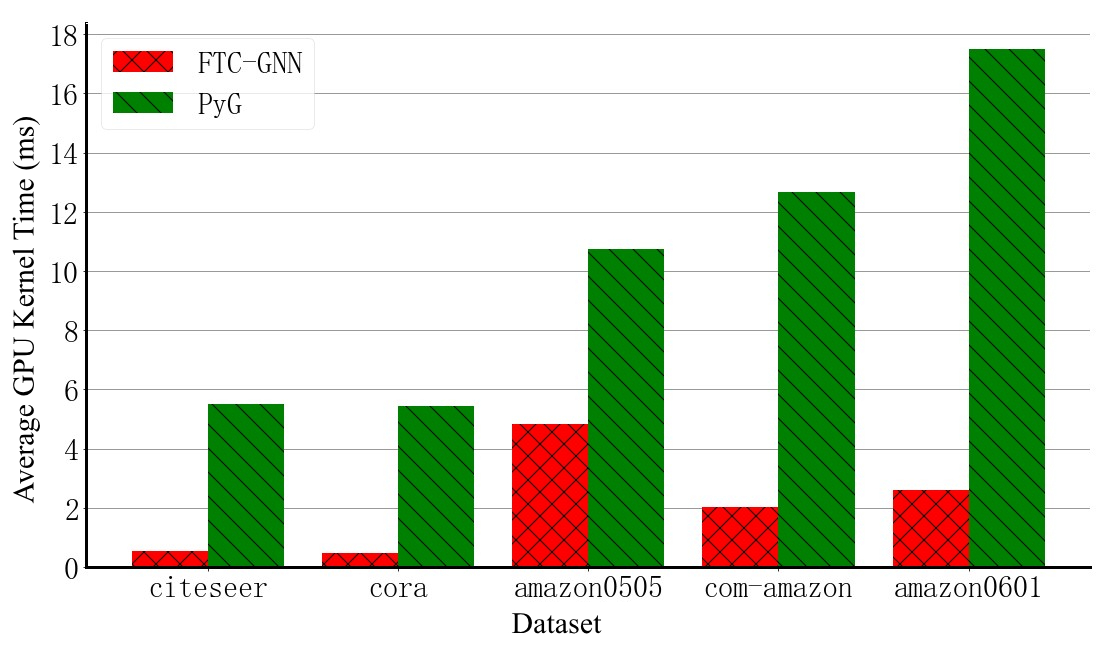}} % Replace with your actual image path
    \caption{Comparison of Average GPU Kernel Time Between FTC-GNN and PyG (AGNN)}
    \label{fig:ftc_pyg_agnn}
\end{figure}

As shown in Figures~\ref{fig:ftc_pyg_gcn} and~\ref{fig:ftc_pyg_agnn}, FTC-GNN achieved an average speedup of 7.10 times for GCN and 2.92 times for AGNN compared to PyG across multiple datasets.

For GCN, the specific average GPU kernel times of FTC-GNN and PyG are shown in Table~\ref{table:ftc_pyg_gcn}. Comparing the average GPU kernel time of FTC-GNN and PyG, FTC-GNN has significantly shorter times across all datasets. For the \texttt{citeseer} and \texttt{cora} datasets, FTC-GNN’s average GPU kernel time is much lower than PyG's. For the \texttt{amazon0505}, \texttt{com-amazon}, and \texttt{amazon0601} datasets, FTC-GNN’s average GPU kernel time is also significantly lower than PyG's.

\begin{table}[h]
    \centering
    \caption{Average GPU Kernel Time of FTC-GNN and PyG (GCN)}
    \label{table:ftc_pyg_gcn}
    \begin{tabular}{|p{1.5cm}|p{3cm}|p{3cm}|}
        \hline
        Dataset & FTC-GNN Average GPU Kernel Time (ms) & PyG Average GPU Kernel Time (ms) \\
        \hline
        citeseer & 0.551 & 5.514 \\
        cora & 0.485 & 5.424 \\
        amazon0505 & 4.846 & 10.752 \\
        com-amazon & 2.025 & 12.670 \\
        amazon0601 & 2.604 & 17.492 \\
        \hline
    \end{tabular}
\end{table}

In summary, FTC-GNN exhibits higher computational efficiency than PyG under the GCN model, with significantly lower average GPU kernel times, indicating that FTC-GNN completes tasks faster, saving computational resources and time.

For AGNN, the specific average GPU kernel times of FTC-GNN and PyG are shown in Table~\ref{table:ftc_pyg_agnn}. Comparing the average GPU kernel time of FTC-GNN and PyG, FTC-GNN’s average GPU kernel time is consistently lower across all datasets. For the \texttt{citeseer} and \texttt{cora} datasets, FTC-GNN’s average GPU kernel time is slightly lower than PyG's. For the \texttt{amazon0505}, \texttt{com-amazon}, and \texttt{amazon0601} datasets, FTC-GNN's average GPU kernel time is also significantly lower than PyG's.

\begin{table}[h]
    \centering
    \caption{Average GPU Kernel Time of FTC-GNN and PyG (AGNN)}
    \label{table:ftc_pyg_agnn}
    \begin{tabular}{|p{1.5cm}|p{3cm}|p{3cm}|}
        \hline
        Dataset & FTC-GNN Average GPU Kernel Time (ms) & PyG Average GPU Kernel Time (ms) \\
        \hline
        citeseer & 1.665 & 5.547 \\
        cora & 1.744 & 5.395 \\
        amazon0505 & 38.762 & 52.209 \\
        com-amazon & 19.179 & 29.043 \\
        amazon0601 & 22.358 & 40.907 \\
        \hline
    \end{tabular}
\end{table}

In summary, under the AGNN model, FTC-GNN shows higher computational efficiency than PyG, with significantly lower average GPU kernel times, indicating that FTC-GNN completes GNN computation tasks faster, saving computational resources and time.

\subsubsection{Comparison with TC-GNN}

\begin{figure}[htbp]
    \centerline{
    \includegraphics[width=0.5\textwidth]{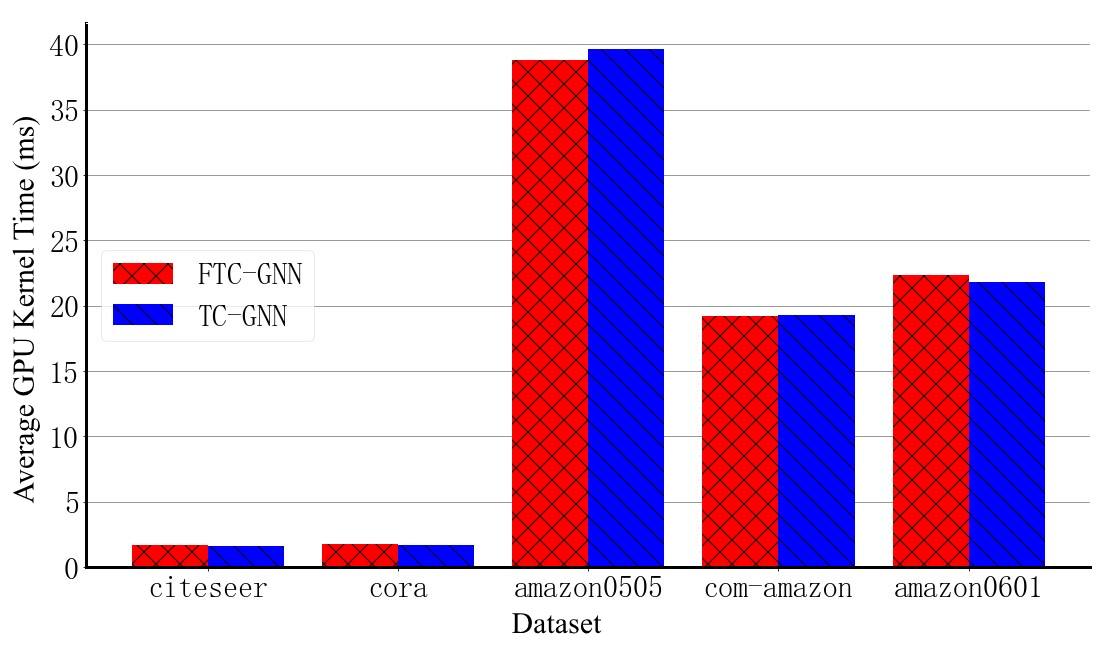}} % Replace with your actual image path
    \caption{Comparison of Average GPU Kernel Time Between FTC-GNN and TC-GNN (GCN)}
    \label{fig:ftc_tc_gcn}
\end{figure}

\begin{figure}[htbp]
    \centerline{
    \includegraphics[width=0.5\textwidth]{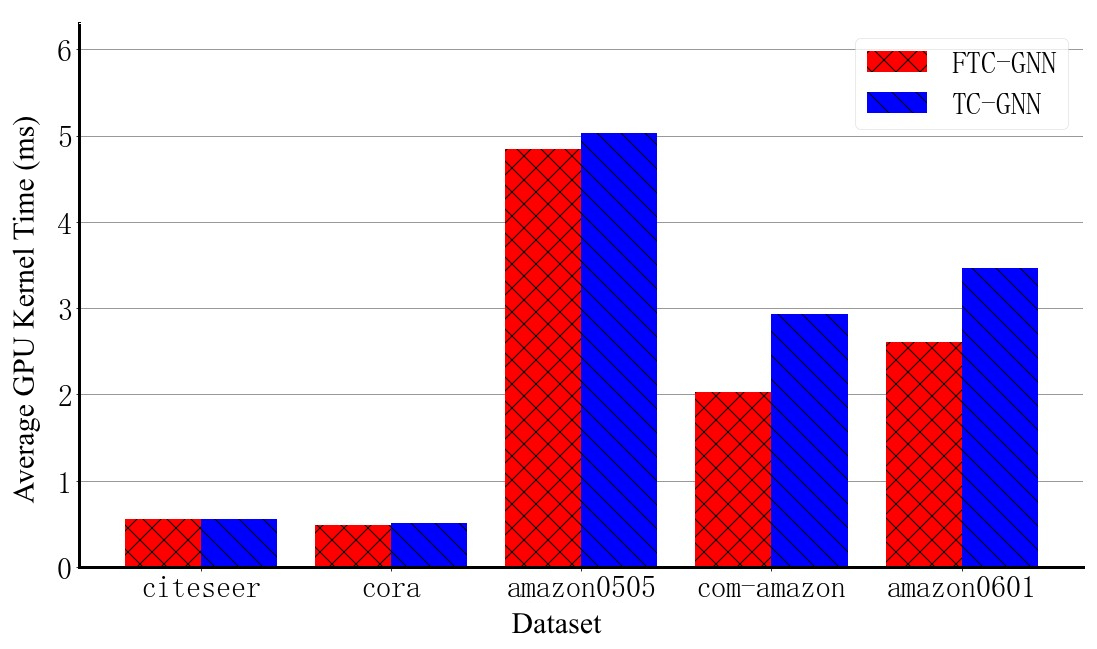}} % Replace with your actual image path
    \caption{Comparison of Average GPU Kernel Time Between FTC-GNN and TC-GNN (AGNN)}
    \label{fig:ftc_tc_agnn}
\end{figure}

Compared to TC-GNN, FTC-GNN achieved an average speedup of 1.17 times for GCN and 1.02 times for AGNN, as shown in Figures~\ref{fig:ftc_tc_gcn} and~\ref{fig:ftc_tc_agnn}. These results indicate that FTC-GNN performs better for GCN acceleration.

For GCN, the specific average GPU kernel times of FTC-GNN and TC-GNN are shown in Table~\ref{table:ftc_tc_gcn}. Comparing the average GPU kernel time of FTC-GNN and TC-GNN, the times are very close across all datasets. In the \texttt{citeseer} dataset, their average GPU kernel times are identical, while in other datasets, TC-GNN's average GPU kernel time is slightly higher than FTC-GNN's.

\begin{table}[htbp]
    \centering
    \caption{Average GPU Kernel Time of FTC-GNN and TC-GNN (GCN)}
    \label{table:ftc_tc_gcn}
    \begin{tabular}{|p{1.5cm}|p{3cm}|p{3cm}|} 
        \hline
        Dataset & FTC-GNN Average GPU Kernel Time (ms) & TC-GNN Average GPU Kernel Time (ms) \\
        \hline
        citeseer & 0.551 & 0.551 \\
        cora & 0.485 & 0.510 \\
        amazon0505 & 4.846 & 5.027 \\
        com-amazon & 2.025 & 2.934 \\
        amazon0601 & 2.604 & 3.470 \\
        \hline
    \end{tabular}
\end{table}

In summary, under the GCN model, FTC-GNN and TC-GNN exhibit similar average GPU kernel times, with close computational efficiency on most datasets.

For AGNN, the specific average GPU kernel times of FTC-GNN and TC-GNN are shown in Table~\ref{table:ftc_tc_agnn}. Comparing the average GPU kernel times, FTC-GNN and TC-GNN have very similar times across all datasets. For the \texttt{citeseer}, \texttt{cora}, and \texttt{com-amazon} datasets, TC-GNN's average GPU kernel time is slightly lower than FTC-GNN's. However, for the \texttt{amazon0505} and \texttt{amazon0601} datasets, FTC-GNN's average GPU kernel time is slightly lower than TC-GNN's.

\begin{table}[h]
    \centering
    \caption{Average GPU Kernel Time of FTC-GNN and TC-GNN (AGNN)}
    \label{table:ftc_tc_agnn}
    \begin{tabular}{|p{1.5cm}|p{3cm}|p{3cm}|}
        \hline
        Dataset & FTC-GNN Average GPU Kernel Time (ms) & TC-GNN Average GPU Kernel Time (ms) \\
        \hline
        citeseer & 1.665 & 1.630 \\
        cora & 1.744 & 1.700 \\
        amazon0505 & 38.762 & 39.614 \\
        com-amazon & 19.179 & 19.273 \\
        amazon0601 & 22.358 & 21.831 \\
        \hline
    \end{tabular}
\end{table}

In summary, under the AGNN model, FTC-GNN and TC-GNN exhibit very similar average GPU kernel times, with minimal differences, indicating similar computational efficiency in GNN computation.

\section{Conclusion and Future Directions}

GNNs are a powerful deep learning model for processing graph-structured data, capable of capturing complex relationships within graphs. However, handling large-scale graphs and efficiently implementing GNNs presents a major challenge. To address this challenge, Tensor Cores have been introduced in this context, providing mixed-precision matrix multiplication and accumulation operations that significantly improve computational efficiency and throughput, thereby accelerating the inference and training processes of GNNs.

First, using Tensor Cores for GNN computation can significantly enhance computational efficiency. Since GNNs require substantial matrix computations at each layer, Tensor Cores allow these operations to be processed in parallel, greatly increasing computation speed. Additionally, Tensor Cores support mixed-precision computation, meaning that computation time and resource requirements can be reduced while maintaining accuracy.

Second, this study designed an efficient algorithmic framework that leverages parallel computing using CUDA Cores and Tensor Cores to improve computational performance. Furthermore, a series of key technical implementations tailored for sparse graph data were proposed, including sparse graph transformation, sparse neighbor aggregation, and sparse edge feature computation. The CUDA function calls were also optimized to further enhance computational efficiency.

However, designing GNN acceleration based on Tensor Cores also presents certain challenges. Although Tensor Cores provide efficient matrix operations, GNN computation involves numerous non-matrix operations, such as node and edge updates. These operations cannot be directly accelerated with Tensor Cores, requiring effective algorithmic designs to handle them.

Additionally, the acceleration design in this study primarily targets specific GNN models and hardware platforms. As GNN models evolve and hardware platforms diversify, ensuring scalability and generalizability of the acceleration design remains a challenge. Future research can address these issues with improvements in the following areas:

\begin{itemize}
    \item \textbf{Optimizing Data Storage and Communication Strategies:} This study's optimization of data storage and communication could be further improved. For instance, exploring more efficient data compression and encoding techniques could reduce storage space requirements and communication overhead. Additionally, studying data distribution and access strategies tailored to specific graph structures may improve memory access performance.
    
    \item \textbf{Model Extension:} The current acceleration scheme is optimized for GCN and AGNN models. Future research could consider extending the scheme to accommodate more diverse types of GNN models, enhancing the generalizability and applicability of the acceleration solution.
    
    \item \textbf{Hardware Optimization:} This study primarily focused on the application of Tensor Cores and CUDA Cores in GNN computation, without addressing other acceleration hardware (such as FPGAs and ASICs). Future work could explore integrating these hardware resources for deeper research.
    
    \item \textbf{Developing an Adaptive GNN Acceleration Framework:} Such a framework could dynamically adjust computational resource allocation based on the characteristics and scale of input graph data, achieving more efficient acceleration.
\end{itemize}

In conclusion, Tensor Core-based acceleration for GNNs will continue to play an important role in advancing GNN technology and its applications. To achieve this goal, further research is needed in data storage optimization, model extension, and hardware optimization to meet evolving computational demands and application scenarios.

\end{document}